\documentclass[10pt, a4paper]{article}

\usepackage[final]{lrec2026} 

\usepackage[main=english,ukrainian]{babel}

\usepackage{graphicx}
\usepackage{tabularx}
\usepackage{array}
\usepackage{hyperref}
\usepackage{natbib}
\usepackage{booktabs}
\usepackage{color, colortbl}
\usepackage{enumitem}
\usepackage{caption}
\usepackage{adjustbox}
\usepackage{amssymb}
\usepackage{utfsym}
\usepackage{mathtools}
\usepackage{amsmath}
\usepackage{inconsolata}
\usepackage{subcaption}
\usepackage{enumitem}
\usepackage{xcolor}
\usepackage[most]{tcolorbox}
\usepackage{textcomp}

\definecolor{lightblue}{RGB}{220, 240, 255}

\title{\textsc{CrossNews-UA}: A Cross-lingual News Semantic Similarity Benchmark for Ukrainian, Polish, Russian, and English}

\name{Daryna Dementieva\textsuperscript{1,2}, Evgeniya Sukhodolskaya\textsuperscript{1*}, Alexander Fraser\textsuperscript{1,2,3}} 

\address{
\textsuperscript{1}Technical University of Munich (TUM), \\
\textsuperscript{2}Munich Center for Machine Learning (MCML), $^{3}$Munich Data Science Institute \\
\href{mailto:daryna.dementieva@tum.de}{\texttt{\small daryna.dementieva@tum.de}}
}

\abstract{
In the era of social networks and rapid misinformation spread, news analysis remains a critical task. Detecting fake news across multiple languages, particularly beyond English, poses significant challenges. Cross-lingual news comparison offers a promising approach to verify information by leveraging external sources in different languages~\cite{DBLP:journals/aim/ChenS24}. However, existing datasets for cross-lingual news analysis~\cite{chen-etal-2022-semeval} were manually curated by journalists and experts, limiting their scalability and adaptability to new languages. In this work, we address this gap by introducing a scalable, explainable \textbf{crowdsourcing pipeline for cross-lingual news similarity} assessment. Using this pipeline, we collected a novel dataset \textsc{CrossNews-UA} of news pairs in \textbf{Ukrainian} as a central language with linguistically and contextually relevant languages---\textbf{Polish}, \textbf{Russian}, and \textbf{English}. Each news pair is annotated for semantic similarity with detailed justifications based on the \textbf{4W} criteria (Who, What, Where, When). We further tested a range of models, from traditional bag-of-words, Transformer-based architectures to large language models (LLMs).  Our results highlight the challenges in multilingual news analysis and offer insights into models performance.
 \\ \newline  }

\begin{document}

\maketitleabstract

\section{Introduction}

\begin{figure}[ht!]
    \centering
    \includegraphics[width=\linewidth]{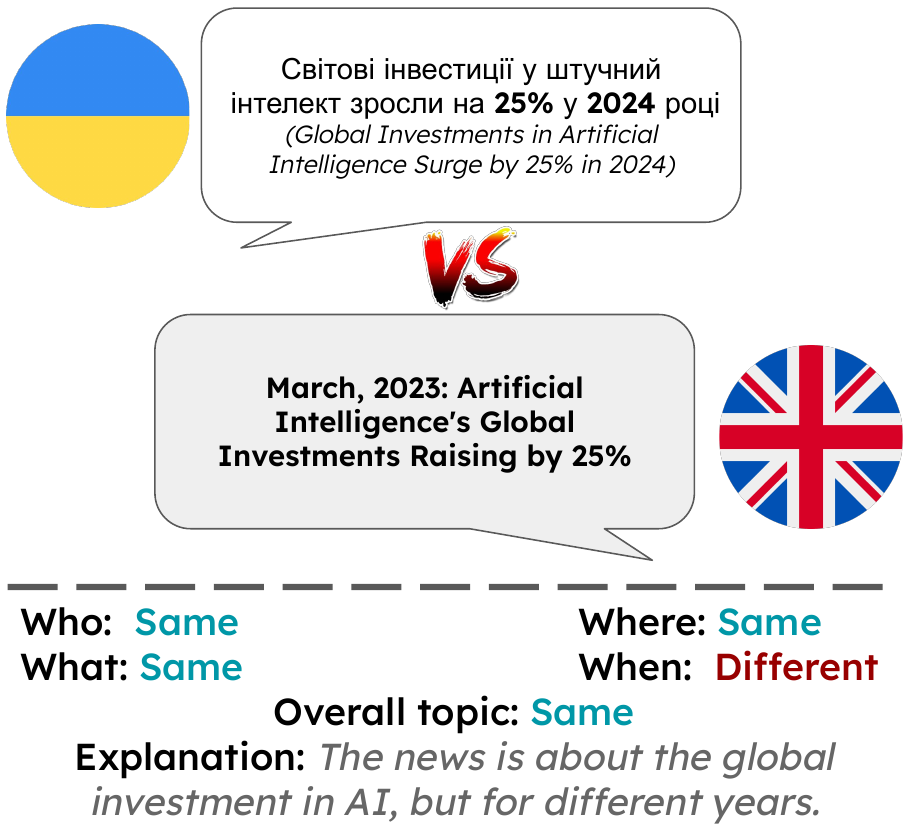}
    \caption{Example of the explainable cross-lingual news comparison obtained in this work for \textsc{CrossNews-UA} corpus.}
    \label{fig:intro}
\end{figure}

The detection of fake news, especially, in a multilingual setup is still a critical challenge. While numerous methods exist for fake news detection---ranging from internal linguistic features~\cite{perez-rosas-etal-2018-automatic,DBLP:journals/snam/AbualigahADAM23} to leveraging external knowledge~\cite{DBLP:conf/clef/GhanemMPR18,chen-etal-2024-complex,DBLP:journals/aim/ChenS24}---external verification from multilingual resources presents a promising direction~\cite{DBLP:journals/jimaging/DementievaKP23}. A key subtask in such a pipeline is cross-lingual news comparison assessing the semantic similarity between news articles in different languages. However, beyond simple similarity scoring, there is a need for explainable data and approaches that provide insights into what main aspects of the content are similar or divergent (e.g., facts, entities, or temporal details, see Figure~\ref{fig:intro}).

Cross-lingual document comparison, in general, is a complex task and lack of resources for many languages. There was introduced a dataset for cross-lingual news comparison for SemEval-2022 Shared Task 8~\citelanguageresource{chen-etal-2022-semeval}. The data collection was done by journalist experts, however, without clear criteria for a regular news reader and difficult to reproduce. While the dataset indeed covered up to 10 languages, it is quite challenging to re-use the proposed expert-dependent pipeline to new languages easily. Given the current global events and the urgent need for reliable information, there is a strong motivation to support languages such as Ukrainian, which are underrepresented in existing resources.

In this paper, we address these gaps with the following contributions:

\begin{itemize}
\item We introduce a novel \textbf{crowdsourcing pipeline} designed to collect high-quality cross-lingual news similarity data, ensuring both scalability and transparency in the annotation process. We designed the pipeline to not just give an overall similarity score, but \textbf{explain} news relations across several dimensions. 
\item We apply this pipeline to create a \textbf{multilingual dataset \textsc{CrossNews-UA} featuring Ukrainian} alongside with \textbf{cross-lingual pairs} with contextually relevant languages---Polish, Russian and English.
\item Using this dataset, we \textbf{test a range of models}, from traditional bag-of-words approaches to Transformer-based ones and LLMs giving the insights on models performance on the multilingual news similarity estimation task.
\end{itemize}

All data, code, and annotation instruction are available online.\footnote{We release code for crowdsourcing project as well as the final dataset: \\ \href{https://github.com/TUM-NLP/crossnews-ua}{https://github.com/TUM-NLP/crossnews-ua} \\ \href{https://huggingface.co/datasets/ukr-detect/crossnews-ua}{https://huggingface.co/datasets/ukr-detect/crossnews-ua}.}



\section{Related Work}



\paragraph{Multilingual Documents Similarities Datasets} The challenge of estimating text similarity---ranging from sentences to paragraphs and full articles---has been widely studied in the research community. For example, \citet{ferrero-etal-2016-multilingual} introduced a multilingual, multi-style, and multi-granularity dataset for cross-lingual textual similarity detection, covering French, English, and Spanish, with data sourced from both manual and machine translations. In SemEval 2017, the Semantic Textual Similarity Multilingual and Cross-lingual Focused Evaluation task was introduced~\citelanguageresource{cer-etal-2017-semeval}, concentrating on sentence-level similarity across five languages. More recently, the SemEval-2022 shared task on multilingual news similarity~\citelanguageresource{chen-etal-2022-semeval} expanded language coverage and evaluated documents based on seven distinct parameters, with datasets curated by domain experts. In recent work, \citet{DBLP:conf/icwsm/ChenSHJG24} expanded the existing framework by incorporating automated news article scraping, further enriching the dataset. Despite these advancements, significant limitations remain: (i) none of the previous studies have introduced a scalable crowdsourcing pipeline adaptable to any language, and (ii) Ukrainian has consistently been excluded from the language coverage.

\paragraph{Multilingual Documents Similarities Estimation Methods} Baseline methods for multilingual text similarity estimation often use sentence encoders like LaBSE~\cite{reimers-gurevych-2020-making} and mBERT~\cite{devlin-etal-2019-bert}, with similarity measured through metrics like cosine similarity. Another approach involves annotating topics from monolingual documents with cross-lingual labels, allowing similarity assessments through multilingual concept hierarchies derived from independently trained models~\cite{DBLP:conf/kcap/Badenes-OlmedoG19}. Additionally, multilingual sequence-to-sequence training with a similarity loss function has been employed to improve cross-lingual document classification, incorporating similarity constraints during training to better identify semantically similar documents across languages~\cite{DBLP:conf/rep4nlp/YuLO18}. For SemEval-2022 Shared Task 8, the top solutions were based on fine-tuning, combining, and stacking various transformer models like DistilBERT, BERT, RoBERTa, and XLM-RoBERTa~\cite{di-giovanni-etal-2022-datascience,kuimov-etal-2022-skoltechnlp,chen-etal-2022-itnlp2022}. LLMs are usually evaluated on the MTEB benchmark~\cite{DBLP:conf/eacl/MuennighoffTMR23} for the STS task, however, this assessment does not cover document-to-document similarity~\cite{DBLP:journals/corr/abs-2303-18223}. \citet{gatto-etal-2023-text} found that generative LLMs outperform encoder-based STS models in assessing complex semantic relationships, but their performance in low-resource languages or domain-specific tasks is still uncertain due to limited annotated data.

\section{\textsc{CrossNews-UA} Corpus Annotation}


We built our pipeline on the experience of the SemEval2022 Task 8~\citelanguageresource{chen-etal-2022-semeval}, adapting it to a broader, more scalable scenario. Rather than relying on academic experts for labelling, we collected data through general crowdsourcing. 




\subsection{News Semantic Similarity Measurement Objective}

\begin{figure*}[ht!]
   \centering
   \includegraphics[width=0.9\textwidth]{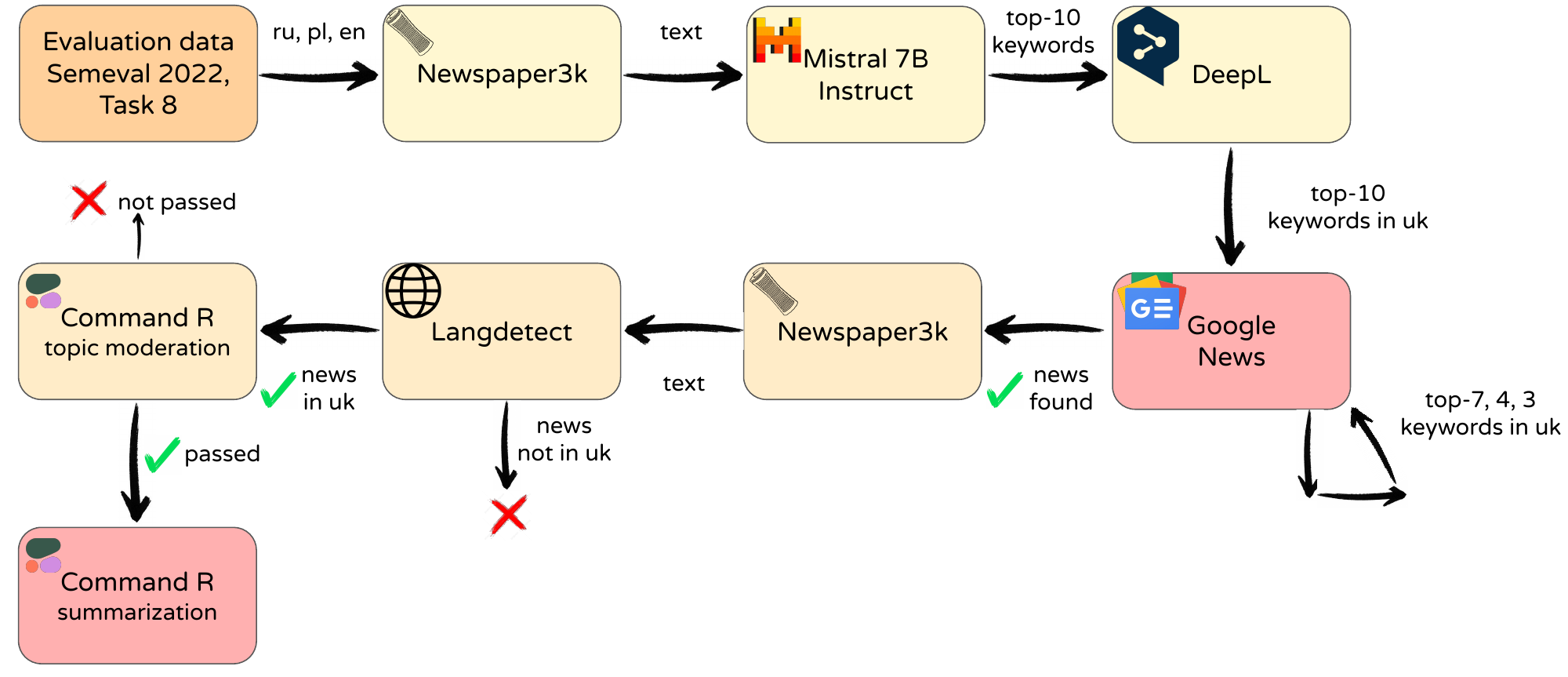}
   \caption{Initial scraping pipeline of news article in Ukrainian.}
   \label{fig:scraping}
\end{figure*}

While crowdsourcing offers scalability, it also presents challenges in ensuring labeling quality and preventing fraud. Clear instructions, an intuitive interface, and a well-structured task design are essential for maintaining data reliability. To simplify the evaluation process, we decomposed the news similarity task using the \textbf{4W model}, derived from the traditional \textit{5W1H} framework commonly used in journalism to summarize key story elements---What, Where, When, Who, Which, and How~\cite{bley1913}. Since comparing news articles aligns closely with evaluating these core elements, we focused on the \textbf{four most critical dimensions}: \textit{Who}, \textit{When}, \textit{Where}, and \textit{What}. Annotators assessed each of these dimensions using \textbf{three straightforward labels}---``Similar'', ``Somewhat related'', and ``Different''---with an optional ''Incomparable'' label ensuring both clarity and ease of use.

To improve the transparency of labeling decisions, annotators were also required to provide brief \textbf{explanations} justifying their similarity assessments, particularly regarding the ``What'' component. These explanations not only clarify annotation choices but also serve as valuable data for fine-tuning NLP models in tasks related to event summarization and semantic similarity.

\subsection{Selecting Multilingual News Articles for Annotation}

The first step was to collect new data pairs that will include Ukrainian news. We developed the several-step pipeline that is schematically illustrated in Figure~\ref{fig:scraping}. For human evaluations of models' performance on Ukrainian text processing, we engaged three fluent Ukrainian annotators, each holding a Master's or PhD in Computer Science.

\subsubsection{Initial News Selection}

To ensure a balanced dataset across similarity levels and languages, we re-used the SemEval 2022 Task 8~\citelanguageresource{chen-etal-2022-semeval} data as a base for the further search. We sampled equal numbers of news pairs from four original categories in this dataset: ``Very Similar'', ``Somewhat Similar'', ``Somewhat Dissimilar'', and ``Very Dissimilar''. Our focus was on only \textbf{related to Ukrainian languages}---Russian, Polish, and English news pairs---given their frequent coverage of events related to Ukraine. The primary objective was to identify a corresponding Ukrainian article for any language in each news pair.

\subsubsection{Scraping News in Ukrainian}
%
\paragraph{Generating Search Request in Ukrainian} To find suitable Ukrainian news for pairs, we needed a concise summary of original articles' key events. Given that data samples' headlines often can be abstract and fail to capture the full content, we extracted the top 10 keywords from each article's content to guide our search. For keyword extraction, we employed the Mistral 7B Instruct model\footnote{\scriptsize{\href{https://huggingface.co/mistralai/Mistral-7B-Instruct-v0.2}{https://huggingface.co/mistralai/Mistral-7B-Instruct-v0.2}}} which demonstrated strong performance in generating relevant keywords in the preliminary experiments.
Then, the keywords were translated into Ukrainian using DeepL\footnote{\scriptsize{\href{https://www.deepl.com}{https://www.deepl.com}}} and used as search queries.


\paragraph{Automated News Retrieval} We sourced news articles using the \texttt{GoogleNews}\footnote{\scriptsize{\href{https://pypi.org/project/GoogleNews}{https://pypi.org/project/GoogleNews}}} Python library. Our initial search queries consisted of the top 10 extracted keywords, joined by spaces. If no relevant articles were retrieved, we incrementally simplified the query by reducing the keywords to the top 7, then 4, and finally the top 3, to broaden the search scope. The articles were scraped with \texttt{Newspaper3k} library.\footnote{\scriptsize{\href{https://newspaper.readthedocs.io}{https://newspaper.readthedocs.io}}} To ensure language accuracy, we used the langdetect\footnote{\scriptsize{\href{https://pypi.org/project/langdetect}{https://pypi.org/project/langdetect}}} library to verify that the articles were indeed in Ukrainian.

\subsection{Annotation Procedure}

We implemented the annotation procedure using the Toloka platform,\footnote{\scriptsize{\href{https://toloka.ai}{https://toloka.ai}}} chosen for its broad pool of speakers in our target languages and its robust quality control features. The following sections detail the entire annotation setup, including data preparation, instructions design, annotators selection, and the quality control measures employed.

To maintain high annotation quality in large-scale crowdsourcing projects, we considered two strategies: (i)~providing \textbf{guiding questions} to direct annotators' attention to key sections of lengthy articles or generating concise, (ii)~\textbf{information-dense summaries}, allowing annotators to quickly grasp the content and selectively verify details in the full text. 



\subsubsection{Preparing the Dataset for Annotation}
\paragraph{Filtering Sensitive Topics} To comply with the Toloka platform's guidelines on restricted content to protect annotators' well-being,\footnote{\scriptsize{\href{https://toloka.ai/docs/guide/unwanted}{https://toloka.ai/docs/guide/unwanted}}} we pre-filtered news article pairs containing sensitive material.
For automated filtering, we employed Cohere's Command-R\footnote{\scriptsize{\href{https://docs.cohere.com/docs/command-r}{https://docs.cohere.com/docs/command-r}}} model, which demonstrated superior performance in detecting sensitive content compared to other open-source models in preliminary experiments.

\paragraph{Preparing Summarization Format} Summarization was employed to assist annotators by structuring content in alignment with the similarity measurement criteria. The summaries emphasized key elements: \textit{What} occurred, \textit{When} and \textit{Where} events took place, and \textit{Who} the main actors were. For this task, we utilized again Command-R model as well.
As Command-R's API supports Retrieval Augmented Generation (RAG),\footnote{\scriptsize{\href{https://docs.cohere.com/docs/retrieval-augmented-generation-rag}{https://docs.cohere.com/docs/retrieval-augmented-generation-rag}}} we implemented two summarization strategies: embedding the full news article directly within the prompt versus supplying it as retrieved content in RAG mode. Upon evaluating both summarization approaches, we observed no consistent difference in overall quality as both methods occasionally produced hallucinations. To mitigate this, we retained both summarization versions and incorporated a feedback mechanism in the annotation interface, enabling annotators to flag cases where hallucinations rendered the task unfeasible. This approach ensured readiness for the subsequent crowdsourcing phase.


\subsubsection{Instructions \& Interface}

To ensure annotators accurately assessed the information in the news pairs, we provided instructions detailing the task requirements, along with clear explanations of the dimensions and classification categories. We designed the interface to provide all needed information together with a comfortable quick look at the news. The instructions that annotators saw in the beginning:

\begin{tcolorbox}[colback=lightblue, colframe=blue!50, width=\linewidth, boxrule=0.5pt, arc=4pt, left=6pt, right=6pt, top=6pt, bottom=6pt]
\footnotesize{
\textbf{Task description}\\
In each task, you will compare two news articles and evaluate their similarity. The news articles in the pair may be in different languages, but this should not
affect your comparison. For example, Wołodymyr Zełenski (Polish) and Volodymyr Zelenskyy (English) are the same actors in an event. For each news article, \textbf{we have provided}:\\
• A \textbf{link} to the news website (be sure to follow it to see the original article, at least the headline and date of publication);\\
• \textbf{Full text} of the article (if you want to check some details right in the task interface);\\
• \textbf{Two summarization} options made by specially trained algorithms (including basic information on four main parameters of events: WHO, WHEN, WHERE
and WHAT happened).

\textbf{Assessing news similarity}\\
You will answer four questions to help us draw a conclusion about the similarity of the news.
}
\end{tcolorbox}

The example of \textit{Who} dimension question:
\begin{tcolorbox}[colback=lightblue, colframe=blue!50, width=\linewidth, boxrule=0.5pt, arc=4pt, left=6pt, right=6pt, top=6pt, bottom=6pt]
\footnotesize{
\textbf{Do the main characters in the news match? (WHO)}\\
The actor of an article can be a specific person (Volodymyr Zelenskyi), an organization/party or any other association of people.\\
• \textbf{Exactly} - the main and secondary actors in the articles are the same. At the same time, it is important to note that one actor can be called differently, for example,
"Volodymyr Zelenskyi" and "President of Ukraine".\\
• \textbf{Partially} (there are overlaps) - Some of the main or secondary actors are the same.\\
• \textbf{No} - The protagonists in the articles do not coincide at all; they have nothing in common. \\
• \textbf{At least one article does not have any actors} - for example, one of the articles states something very vague, e.g. ``the entire population of the Earth'', or the article is a weather forecast.
}
\end{tcolorbox}

\subsection{Annotators Selection}


\paragraph{Language Proficiency} To ensure high-quality annotations, we divided the labelling tasks into separate pools based on language combinations in the news article pairs: Ukrainian-Ukrainian, Ukrainian-Russian, Ukrainian-Polish, and Ukrainian-English. This allowed us to train and evaluate annotators according to their language proficiency. Toloka platform provided pre-filtering mechanisms to select annotators who had passed official language proficiency tests, serving as an initial screening step. In our scenario, we selected annotators that were proficient in both required languages. The language combinations did not affect pricing or quality control procedures, allowing for a consistent evaluation framework across pools.

\paragraph{Training and Exam Phases} Annotators interested in participating first completed an unpaid training phase, where they reviewed detailed instructions and examples with explanations for correct labelling decisions.
Following this, annotators were required to pass a paid by us exam, identical in format to the actual labelling tasks, to demonstrate their understanding of the guidelines. Successful candidates gained access to the main assignments. Exams were manually reviewed by experts within Toloka's platform, using an interface that mirrored the labelling project setup, allowing reviewers to accept or reject submissions based on accuracy.





\subsection{Quality Control}

\begin{figure}[ht!]
    \centering
    \includegraphics[width=0.5\textwidth]{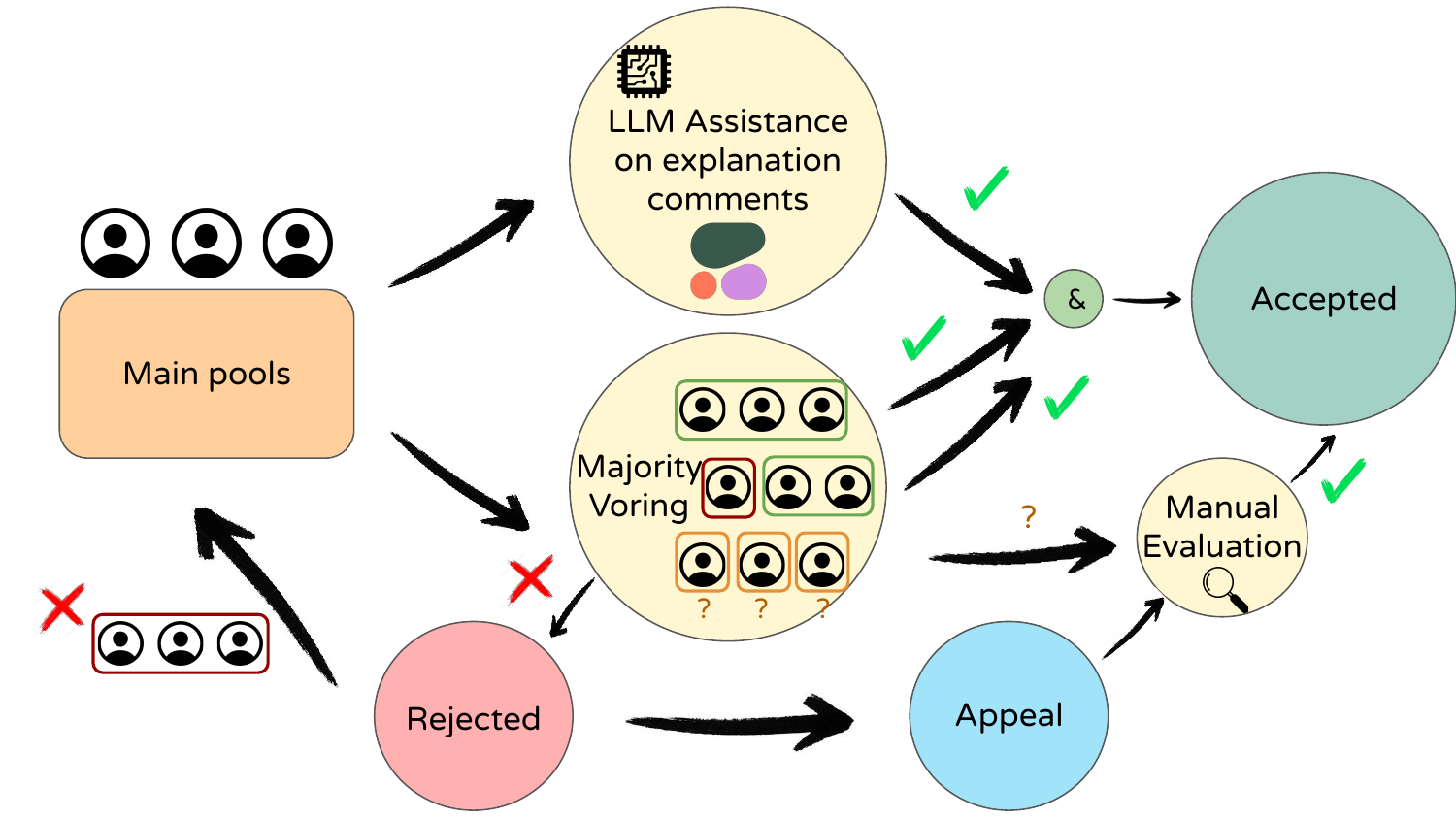}
    \caption{Quality control of labelled assignments overview.}
    \label{fig:qq}
\end{figure}

The quality control pipeline, illustrated in Figure \ref{fig:qq}, was designed with scalability in mind, leveraging automation wherever possible. 

\paragraph{Annotators Overlap} 

To ensure consistent and reliable annotations, each news article pair was assigned to \textbf{three annotators}, enabling the use of automatic \textit{majority voting} for quality control in the classification tasks. This method worked effectively for structured responses to the Who, When, Where, and What questions, each constrained to a limited set of options. However, for the explanatory comments tied to the What question, which involve open-ended text, automatic verification through majority voting was not feasible due to the inherent variability in valid responses.

\paragraph{LLM as an Annotation Quality Control Assistant} To ensure scalable and high-quality annotation, we combined majority voting with the semantic reasoning of Command-R, an LLM used to automatically evaluate annotators' textual explanations and adherence to guidelines. Manual reviews were conducted only in cases of complete annotator disagreement or when rejections were appealed. We fine-tuned prompts by analyzing common annotation errors, especially in event similarity categories like ``Different'' and ``Somewhat Related'', while manually reviewing inconsistencies in ``Similar'' pairs for Who, Where, and When questions.
Command-R demonstrated strong potential as a moderation tool, with a True Positive rate of 96\% and a False Negative rate of 75\%. Additionally, the low rate of appealed rejections (under 10\%) indicated the effectiveness of our quality control process. Tasks rejected were reassigned to highly engaged annotators with strong performance records. This hybrid approach of automated checks and selective manual review successfully balanced quality assurance with scalability.

\begin{figure*}[ht!]
    \begin{subfigure}[b]{0.49\textwidth}
        \centering
        \includegraphics[width=0.6\columnwidth]{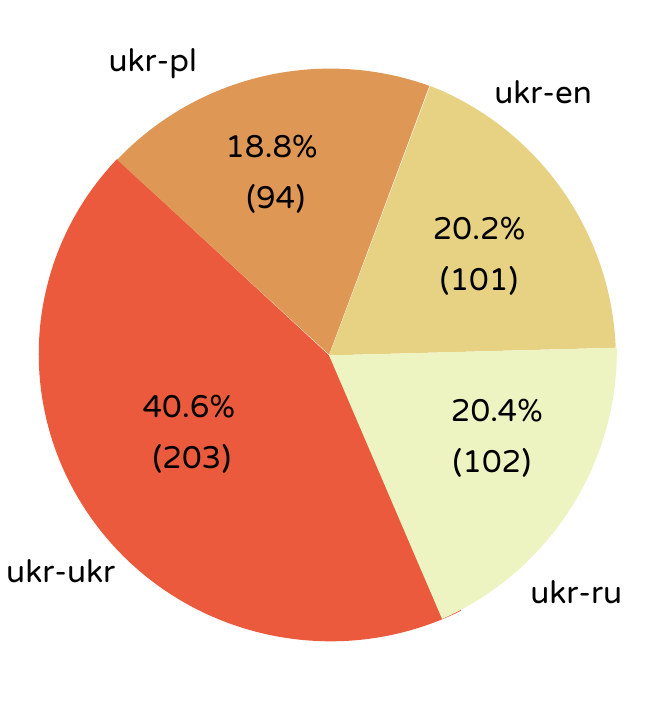}
        \caption{Distribution of Language Pairs}
        \label{fig:pairs_distribution}
    \end{subfigure}
    \hfill
    \begin{subfigure}[b]{0.49\textwidth}
        \centering
        \includegraphics[width=0.7\columnwidth]{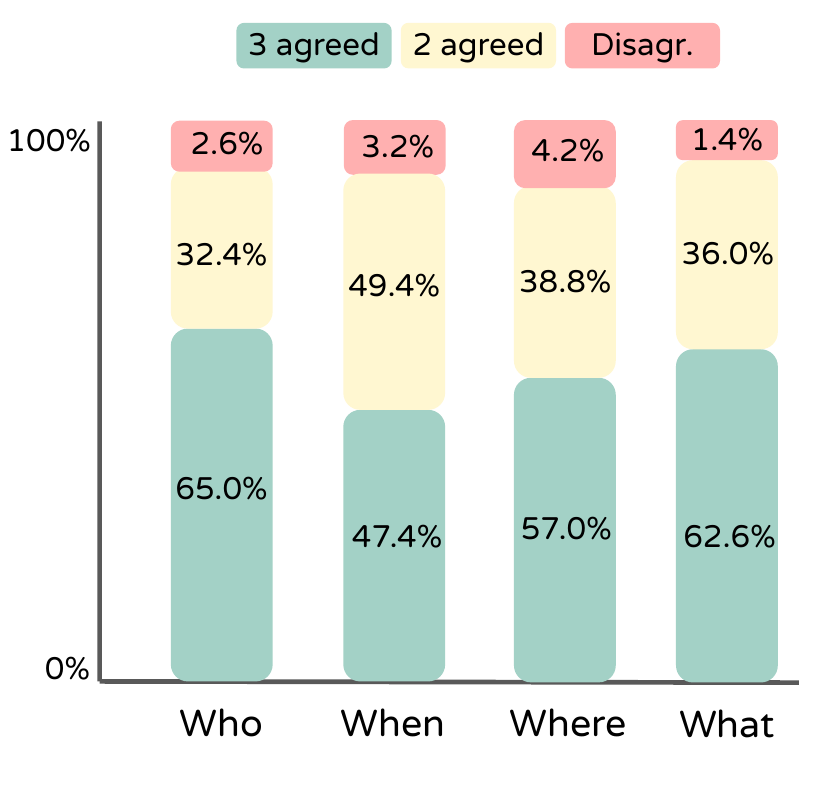}
        \caption{Labels Annotation Agreement per Dimension}
        \label{fig:labels_agreement}
    \end{subfigure}

    \caption{\textsc{CrossNews-UA}: collected final dataset statistics.}
\end{figure*}

\subsubsection{Annotators Well-Being}
\label{sec:annotators_wellbeing}

We aimed to design a fair, transparent, and user-friendly crowdsourcing project, featuring an intuitive interface, comprehensive instructions, and a clear feedback mechanism that allowed annotators to appeal rejected tasks.

\paragraph{Fair Compensation} Payment rates were set to balance grant funding constraints with fair wages, aligning with Ukraine’s minimum hourly wage at the time of labelling (\textbf{1.12 USD/hour}). Annotators received \textbf{0.20 USD} for each accepted exam assignment and \textbf{0.26 USD} for assignments in the main pools (assignment includes only one news pair). Even for tasks rejected by the Majority Vote Quality Control but containing valid comments aligned with instructions, annotators were compensated \textbf{0.15 USD}. The complete project budget is detailed in Table \ref{tab:crowdsourcing_budget}.

\paragraph{Positive Project Ratings} Toloka provided annotators with tools\footnote{\scriptsize{\href{https://toloka.ai/docs/guide/project_rating_stat/}{https://toloka.ai/docs/guide/project\_rating\_stat}}} to rate project fairness in terms of payment, task design, and organizer responsiveness. Our projects received high ratings: \textbf{4.98/5.00} for the Main Project and \textbf{4.95/5.00} for the Training Project.

\subsection{Annotators Statistics}

Out of the $43$ annotators who submitted work in the training phase, $29$ participated in the main pools, with an average of $7$ active users per day. The median age of the annotators was $49$ years, ranging from a minimum of $28$ to a maximum of $74$. The highest number of assignments submitted by a single person was \textbf{138}, while the median number of submissions per person was \textbf{52}. On average, completing a labelling task took \textbf{9.06} minutes. 

\begingroup
\renewcommand{\arraystretch}{1.15}
\begin{table}[hb!]
\centering
\footnotesize
\begin{tabular}{p{3cm}|r|r}
\toprule
\textbf{\shortstack{Category of\\Spending}} & \textbf{\shortstack{Per \\ Assignment (\$)}} & \textbf{Total (\$)} \\
\midrule
Exam assignments & 0.20 & 24.08 \\
Main pools, accepted assignments & 0.20 & 250.04 \\
Bonus, rejected assignments with comments relevant to instructions & 0.15 & 129.95 \\
Additional bonus, accepted assignments & 0.06 & 79.40 \\
\midrule
\textbf{Total} & & \textbf{464.16} \\
\bottomrule
\end{tabular}
\caption{Annotation Expenses.}
\label{tab:crowdsourcing_budget}
\end{table}
\endgroup




\section{Final Dataset}










The resulting dataset contains \textbf{500} pairs of news articles, with the language distribution shown in Figure~\ref{fig:pairs_distribution} with: \texttt{ukr-ukr} $203$ pairs, \texttt{ukr-pl} $94$ pairs, \texttt{ukr-en} $101$ pairs, and \texttt{ukr-ru} $102$ pairs. Additionally, in Figure~\ref{fig:labels_agreement}, we show the amount of annotators agreed on labels per each dimension. The high level of agreement indicates that, despite the complexity of the news semantic similarity task, our new setup effectively streamlined the process for reliable data collection through crowdsourcing.

\begin{figure*}[ht!]
    \begin{subfigure}[b]{0.49\textwidth}
        \centering
        \includegraphics[width=1.1\columnwidth]{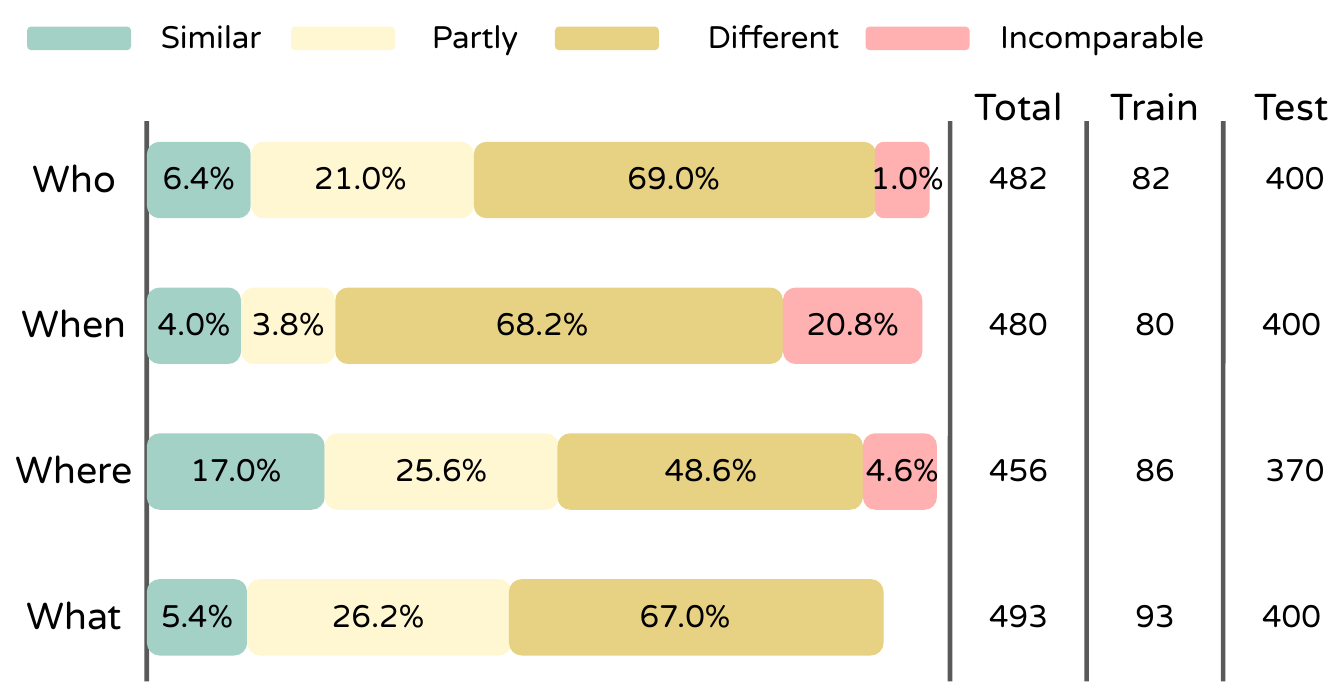}
        \caption{Distribution of Labels per Dimension}
        \label{fig:labels_distribution}
    \end{subfigure}
    \hfill
    \begin{subfigure}[b]{0.49\textwidth}
        \hspace{0.7cm}
        \includegraphics[width=0.95\columnwidth]{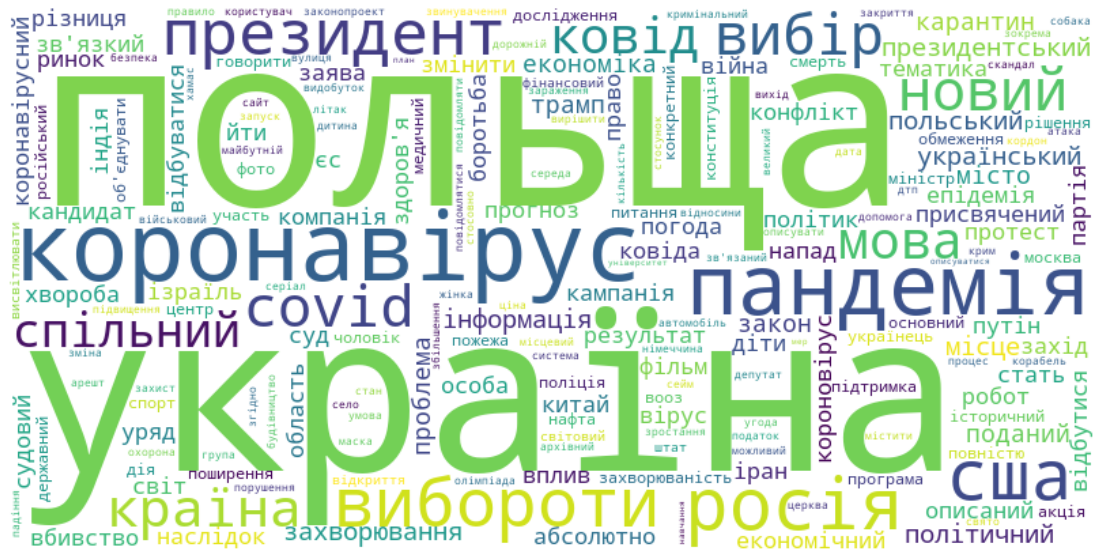}
        \caption{Popular Words in \textit{What} Explanations}
        \label{fig:word_cloud}
    \end{subfigure}

    \caption{Collected dataset labels details.}
\end{figure*}

To establish ground truth labels, we aggregated annotations from overlapping data by \textbf{major voting}, excluding instances of full disagreement within individual categories. The distribution of these aggregated labels across all dimensions is shown in Figure~\ref{fig:labels_distribution}. Notably, the dataset exhibits an imbalance, with the majority of samples labeled as \textit{Different}. For the \textit{When} dimension, a significant portion of samples is marked as \textit{Incomparable}, likely due to challenges in scraping timestamps from original sources and the absence of this information in many articles.

For the \textit{What} dimension, the Ukrainian articles were initially scraped based on pre-existing topics in the dataset, resulting in thematic relationships across the samples without \textit{Incomparable} labels. Figure~\ref{fig:word_cloud} presents an overview of the most common topics mentioned in the \textit{What} explanations, with prominent terms such as \textit{Poland}, \textit{Covid}, \textit{Ukraine}, \textit{elections}, \textit{Russia}, \textit{USA}, \textit{president}, \textit{economics}, \textit{country}, and \textit{borders}. These terms reflect discussions related to politics and economics in Ukraine and neighboring regions.

Finally, we randomly split the dataset into train and test ensuring that both sets contain all labels.




\section{Baselines}
We employed our curated dataset to benchmark multiple models on the Multilingual News Semantic Similarity task. We utilized two types of approaches---\textbf{embeddings-based} and \textbf{prompting-based}.  The task was structured as a classification problem across four dimensions: WHO, WHERE, WHEN, and WHAT. However, the textual explanations provided for the WHAT dimension were excluded from this benchmarking, leaving room for future research focused on explanation generation.

Given the multi-class classification nature of the task, we assessed model performance using standard classification metric \textbf{F1-Score}. To address the labels imbalance, we report \textbf{macro-averaged F1-score}. 




\subsection{Embeddings}

To the best of our knowledge, no existing models are specifically designed for multilingual semantic similarity measurement of news articles that include the Ukrainian language. Thus, we employ multilingual embedding models to generate contextualized embeddings of news article summaries, structured around the Who, Where, When, and What dimensions. The embedding of a text was calculated as the \textbf{mean of its tokens embeddings}. Semantic similarity between news article embeddings is evaluated using \textbf{cosine similarity}. We use a \textbf{decision tree classifier} to find the optimal class decision thresholds of for each model and each question.
The models evaluated include:

\begingroup
\renewcommand{\arraystretch}{1.25}
\begin{table*}[ht!]
\centering
\footnotesize
\begin{tabularx}{\textwidth}
{c|>{\centering\arraybackslash}p{1.1cm}>{\centering\arraybackslash}p{1.1cm}>{\centering\arraybackslash}p{1.1cm}>{\centering\arraybackslash}p{1.1cm}>{\centering\arraybackslash}p{1.1cm}|>{\centering\arraybackslash}p{1.2cm}>{\centering\arraybackslash}p{1.2cm}>{\centering\arraybackslash}p{1.2cm}>{\centering\arraybackslash}p{1.2cm}}
\toprule
\textbf{Category} & \multicolumn{5}{c|}{\textbf{Encoder-based Models}} & \multicolumn{4}{c}{\textbf{LLMs}} \\
\midrule
 & BoW & mBERT & XLM-R & e5 & mT5 & Mistral7B & Mixtral7B & LLaMa3.1 & Aya8B \\
\midrule
Who & 0.35 & 0.30 & 0.31 & \textbf{0.52} & 0.42 & 0.20 & 0.25 & 0.34 & \textbf{0.42}\\
\midrule
When & 0.37 & \textbf{0.49} & 0.13 & 0.28 & 0.27 & 0.12 & 0.35 & \textbf{0.39} & 0.12 \\
\midrule
Where & 0.41 & 0.31 & 0.35 & \textbf{0.49} & 0.28 & 0.26 & 0.34 & \textbf{0.44} & 0.26\\
\midrule
What & 0.50 & 0.17 & 0.15 & \textbf{0.58} & 0.37 & 0.35 & \textbf{0.58} & 0.51 & 0.35\\
\bottomrule
\end{tabularx}
\caption{Macro-averaged F1-score of baselines tested on our proposed \textsc{CrossNews-UA} dataset. The results in \textbf{bold} denotes the best scores per dimension per model type.}
\label{all_metrics}
\end{table*}
\endgroup

\paragraph{Bag-of-Words} Due to its implementation design, the Bag-of-Words model cannot be directly applied for cross-lingual similarity measurement. Therefore, we translated the Russian, Polish, and English summarization parts into Ukrainian using DeepL. Then, we transformed texts into vectors using \texttt{CountVectorizer}. 

\paragraph{BERT} We used the multilingual version of BERT\footnote{\scriptsize{\href{https://huggingface.co/google-bert/bert-base-multilingual-cased}{https://huggingface.co/google-bert/bert-base-multilingual-cased}}}~\cite{DBLP:journals/corr/abs-1810-04805} as it contains all target languages in the pre-trained data.

\paragraph{XLM-RoBERTa} As an extension of BERT-alike models, we used XLM-Roberta-base\footnote{\scriptsize{\href{https://huggingface.co/FacebookAI/xlm-roberta-base}{https://huggingface.co/FacebookAI/xlm-roberta-base}}}~\cite{DBLP:journals/corr/abs-1911-02116} as it shown previously promising results in Ukrainian texts classification~\cite{dementieva-etal-2025-cross}.

\paragraph{mT5} We also selected the encoder part of mT5-small model\footnote{\scriptsize{\href{https://huggingface.co/google/mt5-small}{https://huggingface.co/google/mt5-small}}}~\cite{DBLP:conf/naacl/XueCRKASBR21} to get embeddings of our texts.

\paragraph{Multilingual E5-large} Finally, we utilized the multilingual e5-large model\footnote{\scriptsize{\href{https://huggingface.co/intfloat/multilingual-e5-large}{https://huggingface.co/intfloat/multilingual-e5-large}}}~\cite{DBLP:journals/corr/abs-2402-05672} that already showed promising results in various multilingual text similarity and extraction tasks.

\subsection{LLMs} 
To test models based on another methodology, we also tried out various LLMs on our benchmark dataset transforming our classification task into the \texttt{text-to-text} generation one. While Ukrainian and other target languages are not always explicitly present in the pre-training data reports, the emerging abilities of LLMs already showed promising results in handling new languages~\cite{DBLP:journals/tmlr/WeiTBRZBYBZMCHVLDF22} including Ukrainian~\cite{dementieva-etal-2025-cross}. With the prompt mentioned in Appendix~\ref{sec:app_llm_similarity_prompt}, we included the following models in evaluation:

\paragraph{Mistral} We used several version of Mistral-family models~\cite{Jiang2023Mistral7}---Mistral-7B-Instruct\footnote{\scriptsize{\href{https://huggingface.co/mistralai/Mistral-7B-Instruct-v0.3}{https://huggingface.co/mistralai/Mistral-7B-Instruct-v0.3}}}
and Mixtral-8x7B-Instruct.\footnote{\scriptsize{\href{https://huggingface.co/mistralai/Mixtral-8x7B-Instruct-v0.1}{https://huggingface.co/mistralai/Mixtral-8x7B-Instruct-v0.1}}} The models cards do not mention explicitly Ukrainian and other languages, however Mistral showed promising results in Ukrainian texts classification tasks~\cite{dementieva-etal-2025-cross}.

\paragraph{Llama3} Also, we tested the Llama-3.1-8B-Instruct model\footnote{\scriptsize{\href{https://huggingface.co/meta-llama/Meta-Llama-3.1-8B-Instruct}{https://huggingface.co/meta-llama/Meta-Llama-3.1-8B-Instruct}}}~\cite{llama3modelcard}. The model card as well does not stated Ukrainian explicitly, however, encourages research in usage of the model in various multilingual tasks.

\paragraph{Aya-23 8B} Finally, we chose the Aya-23-8B\footnote{\scriptsize{\href{https://huggingface.co/CohereForAI/aya-23-8B}{https://huggingface.co/CohereForAI/aya-23-8B}}} model~\cite{DBLP:journals/corr/abs-2405-15032}. All target languages were explicitly present in its pre-training data.

\section{Results}

The final multilingual results are presented in Table~\ref{all_metrics}.

The results reveal several key insights. First, \textbf{When} emerged as the most challenging category for all models. This is understandable, as temporal information can be derived from both the article's content and its publication date. However, since annotators were instructed not to rely on publication dates as the primary source, this information was excluded from model inputs. For prompt-based models, incorporating publication dates into the instructions might have improved performance. 

Second, among encoder models, \textbf{e5-large} achieved the highest performance. This is consistent with its status as a more advanced model widely adopted in modern NLP tasks like Information Retrieval, indicating its strong potential as a base model for future applications in news semantic similarity measurement. Interestingly, a simple \textbf{Bag-of-Words model} outperformed several multilingual, semantically aware encoders still prevalent in the field. This suggests that domain- and language-agnostic approaches like Bag-of-Words remain effective, particularly in low-resource or complex domains.

Lastly, among decoder-only models, \textbf{LLaMa 3.1 8B Instruct} delivered the best performance, which was unexpected given that \textbf{Aya-23 8B} was designed with multilingual tasks in mind. Despite this, Llama-3.1 8B Instruct currently appears to be the most promising candidate for future fine-tuning in the context of news semantic similarity tasks.

Additionally, we provide a closer look into the results of one of the best performing baselines \textbf{e5-large embeddings} per language pair in Table~\ref{tab:e5_results}. Overall, we can see the consistent performance---the results for the monolingual \texttt{ukr-ukr} pair are usually higher than for cross-lingual ones. Especially, for the \texttt{Who} dimension which can be explained by the difference in named entities---the mention of \textit{\selectlanguage{ukrainian}{наш президент}} \textit{(our President)} may also appear as \textit{the current President of Ukraine} (both meaning, for instance, \textit{President Volodymyr Zelenskyy}), highlighting the need to enhance the pipeline with time-aware entity linking grounded in dynamic knowledge bases and highlights the challenge of cross-lingual news comparison.

\begingroup
\renewcommand{\arraystretch}{1.25}
\begin{table}[ht!]
\centering
\footnotesize
\begin{tabular}{c|cccc}
\toprule
\textbf{Category} & \texttt{ukr-ukr} & \texttt{ukr-en} & \texttt{ukr-pl} & \texttt{ukr-ru} \\
\midrule
Who & 0.63 & 0.57 & 0.49 & 0.49 \\
When & 0.30 & 0.27 & 0.33 & 0.27 \\
Where & 0.46 & 0.47 & 0.51 & 0.50 \\
What & 0.61 & 0.59 & 0.45 & 0.55 \\
\bottomrule
\end{tabular}
\caption{The detailed results for \textbf{e5-large embeddings} baseline per each language pair and dimension (macro-averaged F1-score).}
\label{tab:e5_results}
\end{table}

\section{Conclusion}
In this work, we addressed the gap in multilingual news similarity datasets by introducing a novel crowdsourcing pipeline for explainable data collection across underrepresented languages. We provided a comprehensive overview of each stage of the pipeline, including data scraping and preparation, annotation platform setup, instruction and interface development, annotators selection, and quality control processes. To enhance scalability and efficiency, we incorporated automation where possible, utilizing LLMs as quality control assistants, thereby demonstrating the pipeline's adaptability to any digitalized language. Through this approach, we collected \textsc{CrossNews-UA} that consists of $500$ news article pairs in various language pairs combinations including Ukrainian, Polish, Russian, and English. Based on the collected data, we tested a range of embedding-based and LLM-based baselines offering valuable insights into the challenges of the news semantic similarity task for these languages. We believe our dataset will contribute significantly to advancing explainable news comparison and trustworthy information detection.

\section*{Limitations}
\paragraph{Presence of Underrepresented Languages} While our dataset significantly addresses the underrepresentation of Ukrainian in the news semantic similarity task, there remains substantial room for enhancement. We provide detailed budget calculations to guide future data collection efforts, which could include expanding the dataset to cover additional languages, particularly other Slavic languages, to improve cross-linguistic representation. 

\paragraph{Fixed Timestamp of Collection} Furthermore, the current dataset is based on news topics from 2022; incorporating more recent events would enhance its relevance and applicability. Nonetheless, we believe that the news comparison pipelines developed using our data are adaptable and can be effectively transferred to new domains. 

\paragraph{Limited Dataset Size} Our dataset comprises 500 news article pairs with a relatively balanced representation of languages, though label distribution across dimensions remains uneven. Despite this, some of the tested out-of-the-box models demonstrated promising performance. We hope this dataset will serve as a valuable benchmark for future model evaluations involving Ukrainian. Additionally, exploring various data augmentation techniques presents a promising direction for further research.

\paragraph{Annotation Platform} We hope that provided instructions and annotation details make it possible to run the annotation pipeline at any annotation platform or interface without limiting to Toloka.

\paragraph{Baselines} In terms of model benchmarking, we employed straightforward baseline approaches, leaving opportunities for more advanced experiments. Future work could involve end-to-end fine-tuning of Transformer-based models or assembling more complex architectures to improve performance. 

\paragraph{LLMs Choice} Additionally, while we focused on open-source LLMs for prompting, proprietary models were not included in our evaluation, representing another avenue for exploration. 

\paragraph{Multilingual and Cross-lingual Experiments} In our experiments, we perform experiments with only multilingual setup. However, more nuanced insights could be gained by developing and evaluating language technologies tailored to specific language pairs.

\paragraph{Explainability Generation Experiments} Finally, further research could delve into explanation generation techniques aligning with our dataset to enhance the interpretability and explainability of model outputs.

\section*{Ethics Statement}
Our pipeline and dataset were developed with two primary objectives in mind: first, to enhance document representation and comparison for underrepresented languages, particularly Ukrainian; and second, to support research on robust, explainable methods for trustworthy news detection, especially concerning events in Ukraine and neighboring countries. We believe this benchmark will contribute to advancing the field of multilingual news analysis and foster research in more robust and explainable trustworthy news detection covering events in Ukraine and neighbouring countries.

Throughout the data collection process, we adhered to Toloka's ethical guidelines, placing a strong emphasis on annotator well-being as described in Section~\ref{sec:annotators_wellbeing}. We ensured fair compensation, aligned with or exceeding the minimum wage in Ukraine, and maintained open communication with annotators to address any concerns or questions promptly. Regarding data sharing, we will follow the precedent set by SemEval 2022 Task 8~\cite{chen-etal-2022-semeval} by releasing source links to the news articles rather than the full scraped texts, ensuring compliance with copyright regulations.

\section*{Acknowledgments}
We would like to express our gratitude to Toloka.ai platform for their research grant for data annotation. The work was funded/co-funded by the European Union (ERC, EPICAL, 101141712). Views and opinions expressed are however those of the author(s) only and do not necessarily reflect those of the European Union or the European Research Council. Neither the European Union nor the granting authority can be held responsible for them.

\section{Bibliographical References}\label{sec:reference}

\bibliographystyle{lrec2026-natbib}
\bibliography{lrec2026-example,anthology,custom}

\section{Language Resource References}
\label{lr:ref}
\bibliographystylelanguageresource{lrec2026-natbib}
\bibliographylanguageresource{languageresource}

\end{document}